%% file: aaai25.tex
\documentclass[letterpaper]{article} 
\usepackage{aaai25}  
\usepackage{times}  
\usepackage{helvet}  
\usepackage{courier}  
\usepackage[hyphens]{url}  
\usepackage{graphicx} 
\usepackage{natbib}  
\usepackage{caption} 
\usepackage{float}
\usepackage{multirow}
\usepackage{algorithm}
\usepackage{algpseudocode}
\usepackage{amsmath}
\usepackage{subcaption}

\usepackage{newfloat}
\usepackage{listings}
\DeclareCaptionStyle{ruled}{labelfont=normalfont,labelsep=colon,strut=off} 
\floatstyle{ruled}
\newfloat{listing}{tb}{lst}{}
\floatname{listing}{Listing}
\newcommand{\sys}{ATLASv2}

\title{\sys{}: LLM-Guided Adaptive Landmark Acquisition \\ and Navigation on the Edge}
\author {
    Mikolaj Walczak, 
    Uttej Kallakuri, 
    Tinoosh Mohsenin
}
\affiliations {
    Johns Hopkins University\\ 
    mwalcza1@jh.edu, 
    ukallak1@jh.edu, 
    tinoosh@jhu.edu
}

\usepackage{bibentry}

\begin{document}

\maketitle

%

\input{sections/abstract}
\input{sections/introduction}
\input{sections/related_works}
\input{sections/proposed_approach}
\input{sections/results}
\input{sections/conclusion}
\input{sections/acknowledgement}
\small
\bibliography{aaai25}

\end{document}

%% file: sections/abstract.tex
\begin{abstract}
Autonomous systems deployed on edge devices face significant challenges, including resource constraints, real-time processing demands, and adapting to dynamic environments. This work introduces \sys{}, a novel system that integrates a fine-tuned TinyLLM, real-time object detection, and efficient path planning to enable hierarchical, multi-task navigation and manipulation all on the edge device, Jetson Nano. \sys{} dynamically expands its navigable landmarks by detecting and localizing objects in the environment which are saved to its internal knowledge base to be used for future task execution. We evaluate \sys{} in real-world environments, including a handcrafted home and office setting constructed with diverse objects and landmarks. Results show that \sys{} effectively interprets natural language instructions, decomposes them into low-level actions, and executes tasks with high success rates. By leveraging generative AI in a fully on-board framework, \sys{} achieves optimized resource utilization with minimal prompting latency and power consumption, bridging the gap between simulated environments and real-world applications.    
\end{abstract}

%% file: sections/introduction.tex
\section{Introduction}

The growing demand for autonomous systems capable of navigating and manipulating objects in real-world environments has driven advancements in artificial intelligence, robotics, and edge computing \cite{10.1007/s10586-024-04686-y, 10.1109/LES.2024.3446948, prakash2024using, mazumder2023reg}. These systems must operate with minimal latency, high efficiency, and robust adaptability in dynamic, unstructured environments. This paper explores deploying generative AI on the edge \cite{ragib2024vitreg}, using large language models (LLMs) as dynamic planners for hierarchical, multi-task autonomous navigation.


\textbf{Edge computing}, the paradigm of processing data locally on devices rather than relying on cloud infrastructure, plays a pivotal role in this work. Processing on the edge offers several advantages, including enhanced privacy and security, and improved reliability in environments with limited or no internet connectivity. In addition, edge deployment minimizes the dependency on external servers, ensuring consistent performance even in challenging operational scenarios \cite{SINGH202371}.

This paper specifically focuses on the deployment of generative AI on the edge to enable natural language-driven navigation and task execution. LLMs are used not only to interpret high-level natural language instructions but also to dynamically generate actionable plans for autonomous robots \cite{pmlr-v205-shah23b}. We integrate navigation and manipulation tasks, allowing the system to adapt seamlessly to new environments. By dynamically scheduling processes like LLM execution and object detection, the proposed system optimizes onboard power consumption and latency, ensuring efficient operation on resource-constrained hardware.

To provide the Large Language Model (LLM) with information about landmarks in the environment we integrate an onboard object detector. The objects detected by the detector expand the number navigable landmarks enabling the LLM to use the new landmarks for future tasks.


This work bridges the gap between simulated environments and real-world deployment by presenting a resource-efficient, edge-based system capable of performing complex, hierarchical tasks with high-level reasoning and adaptability. The findings highlight the potential of generative AI at the edge for advancing intelligent planning with autonomous systems in diverse and challenging settings. The contributions of this paper are summarized as follows:

\begin{itemize}
    \item A fully on-board solution and system architecture that integrates a path planning module, object detection, and a large language model for navigation and object manipulation tasks.
    \item A novel method for continuously expanding the set of skills and navigable landmarks by detecting objects in the environment and incorporating them into the system's knowledge base with the assistance of the LLM.
    \item Minimizing latency and onboard power consumption through dynamic scheduling of processes, such as LLM execution and object detection.
\end{itemize}

%% file: sections/related_works.tex
\section{Related Work}
The integration of LLMs into robotics is a rapidly growing area, aiming to enable embodied AI that can interpret and execute natural language way-finding instructions in dynamic environments. Many of these works target a fully simulated environment or when deployed use very large and intelligent cloud-based solutions or high power devices \cite{dorbala2024llmsgeneratehumanlikewayfinding,rajvanshi2024saynavgroundinglargelanguage, 10543121, kallakuri2024atlas, 10734363}. For instance, \cite{song2023llmplannerfewshotgroundedplanning} introduced LLM-Planner, a system that leverages few-shot learning to allow embodied agents to interpret and execute high-level natural language instructions. This approach improves adaptability to diverse tasks with minimal prior examples, enhancing their versatility in dynamic settings. LLM-Planner demonstrates the deployment of a robotic agent equipped with a low-level planner in a virtual environment, integrating a pre-trained BERT-base-uncased model \cite{DBLP:journals/corr/abs-1810-04805} with the ChatGPT API, enabling embodied navigation based on high-level plans.

In this work we deploy an LLM for high-level planning on the edge device Jetson Nano, which presents significant challenges with identifying compact models that are able to operate within strict resource constraints. Additionally, many open-source models are trained on generalized datasets, which can result in outputs lacking the precision and focus necessary for effective high-level task decomposition. This misalignment can lead to inefficiencies, consuming resources on decoding and generating responses that do not meet task-specific requirements. To address these challenges, \cite{chen2024octoplanner} proposed the "Octo-planner," a framework that separates planning and action components, optimized for edge devices. The Octo-planner fine-tunes LLMs to improve success rates and efficiency, tackling issues such as context length and computational costs. By utilizing techniques like multi-LoRA training, this framework supports multi-domain queries with reduced resource demands, making it suitable for mobile devices. The Octo-planner demonstrates significant potential for scalable, real-time, and privacy-preserving AI applications on resource-constrained platforms.

This work focuses on the deployment of a robotic agent within a controlled environment that integrates a high-level LLM planner~\cite{bharat2024using}, a low-level motion planner, and an object detector, on the Jetson Nano edge device. This paper provides practical insights into the deployment of integrated onboard systems, inspired by frameworks such as \cite{kallakuri2024atlas}, \cite{micro2023-mozhgan} and \cite{song2023llmplannerfewshotgroundedplanning}. In integrating the high-level LLM planner, we adopt an approach similar to the Octo-Planner framework \cite{chen2024octoplanner}, fine-tuning a compact language model to enhance system reliability, generate task-specific responses, and minimize latency. A significant challenge with the Octo-Planner model is that with 2 billion parameters, even in its open-source form and after quantization, it occupies approximately 2 GB of memory. This poses a risk of exceeding the Jetson Nano's limited 4 GB memory capacity when coupled with an object detector and motion planner. To address this limitation, we incorporate the state-of-the-art TinyLLaMAv1.1 model \cite{zhang2024tinyllama}, an open-source, compact language model with 1.1 billion parameters designed for efficiency and accessibility. TinyLLaMA achieves competitive performance while significantly reducing computational overhead, making it well-suited for resource-constrained environments. By fine-tuning and quantizing the model, we further optimize its performance for real world deployment.

%% file: sections/proposed_approach.tex
\begin{algorithm}[bt]
\caption{The integration of \sys{} components is formalized in the algorithm below, which outlines the dynamic interaction between the navigation, object detection, object interaction and KB building processes.}
\begin{algorithmic}[1]
\Function{main}{}
    \State \Call{startkbprocess}{$initialKB$}
    \While{True}
        \State Wait for: $prompt$
        \State $subtasks \gets LLM(prompt)$
        \State \Call{exectasks}{$subtasks$}
    \EndWhile
\EndFunction

\Function{exectasks}{$subtasks$}
    \ForAll{$action,obj \in subtasks$}
    \If{$obj \notin KB$}
    \State \Return $fail$
    
    \ElsIf{$action$ is $navigate$}
    \State $coo \gets \text{KB}(obj)$
    \State $ros\_navigate(coo)$
    
    \Else    
    \State$moveArm(obj, action)$
    \EndIf
    \EndFor
    \State $spin()$
    \State \Return $success$
\EndFunction

\Function{startkbprocess}{$initialKB$}
    \State $KB \gets initialKB$
    \While{True}
        \State $dets \gets \text{detectAndFilterObjs}()$
        \ForAll{$obj \in dets$}
            \State $objPos, objOrient \gets \text{getCurPos}()$
            \State $KB[obj] \gets objPos, objOrient$
        \EndFor
    \EndWhile
\EndFunction

\end{algorithmic}
\label{algo:1}
\end{algorithm}

\begin{figure*} [hbt!] \centerline{\includegraphics[width=0.93\textwidth]{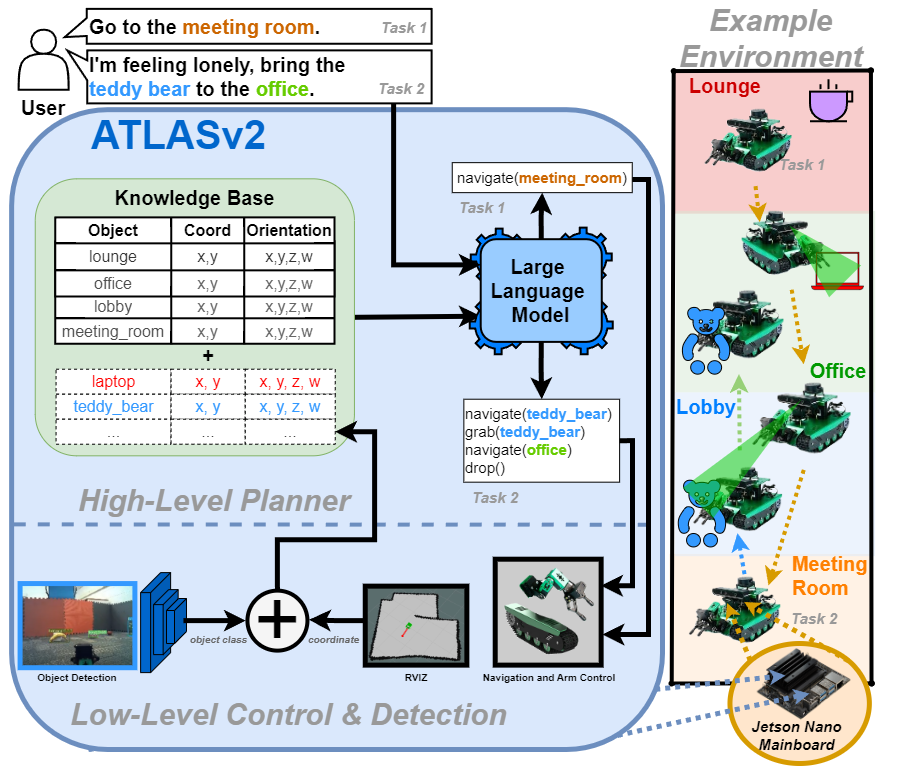}}
    \caption{Block diagram showing how \sys{} is used given two sample prompts provided by the user. The system begins with a prompt received from the user (Task 1) for which the knowledge base is appended to provide the available landmarks. The LLM then generates a plan (if available) that is executed in sequence by the navigation and manipulation package. During execution if a new object of interest is detected (eg. laptop and teddy bear), the object and its location, visualized using RVIZ \cite{10.1007/s11235-015-0034-5} , is appended to the expanding knowledge base. The user can then use the newly detected objects for future tasks (Task 2). The full system is deployed into the Jetson Nano edge device and receives prompts remotely from the user.}
    \vspace{-15pt}
    \label{fig:plan_flow}
\end{figure*}

\begin{figure*} [bt] 
    \centering
    
    \begin{subfigure}[b]{0.25\textwidth}
        \centering
        \includegraphics[width=\linewidth]{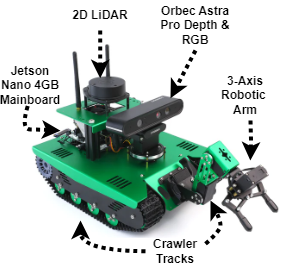}
        \caption{}
    \end{subfigure}
    \hfill
    \begin{subfigure}[b]{0.37\textwidth}
        \centering
        \includegraphics[width=\textwidth]{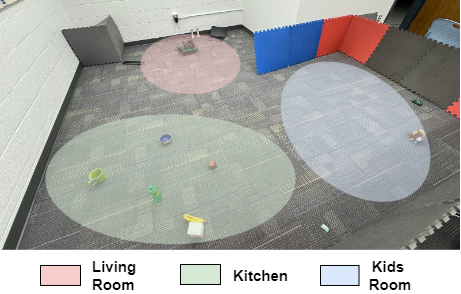}
        \caption{}
    \end{subfigure}
    \hfill    
    \begin{subfigure}[b]{0.37\textwidth}
        \centering
        \includegraphics[width=\textwidth]{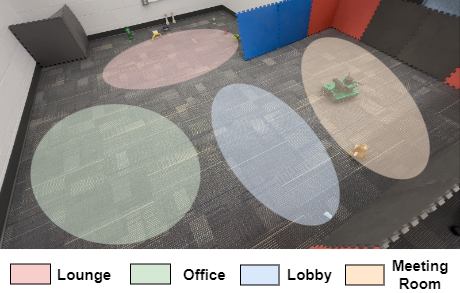}
        \caption{}
    \end{subfigure}
    \caption{The robotic agent used for deployment of \sys{} (a) Yahboom Transbot which contains a 2D lidar for mapping, an Orbbec Astra Pro camera for depth and RGB input, a 3-Axis robotic arm for manipulating objects, crawler tracks for moving around the environment, and a Jetson Nano. Additionally, the configurations of the real-world environment used for system evaluation including (b) The home emulated environment, which includes a living room, kitchen, and kids room, each populated with objects typical of a residential setting and (c) the office emulated environment, comprising of a lounge, office, lobby, and meeting room, with items representative of a professional workplace.}
    
    \label{fig:environment}
\end{figure*}

\section{Proposed Approach}
The proposed system integrates key components to enable hierarchical multi-task autonomous navigation on the Jetson Nano edge device. This architecture combines efficient path planning, real-time object detection, and high-level reasoning with an LLM, all while operating within the strict computational and energy constraints. The approach focuses on leveraging generative AI at the edge for dynamic task planning and knowledge enrichment.


At the core of the system is the interaction between the path planning module, object detection module, and LLM-based planner. These components enable the agent to interpret high-level natural language instructions, plan and execute navigation tasks, and adapt to new environments by expanding its knowledge base (KB). The high-level interaction of these components is outlined in Algorithm \ref{algo:1},

When using the system a natural language prompt describing the desired task is provided. The LLM planner interprets the input and generates a structured plan. This plan decomposes the task into smaller steps, which include navigating to specific landmarks, and grabbing and dropping an object. During task execution newly detected objects and their locations are added to the internal knowledge base for future tasks. This process is outlined in Figure \ref{fig:plan_flow} showing the interaction of the elements of \sys{} when provided two sample prompts in an office emulated environment. All components, including the object detector and LLM planner are deployed directly on the Jetson Nano to ensure reliable operation in areas devoid of a network connection, essential for real-time decision-making. The core components of \sys{} are outlined as follows:

\textbf{Path Planning and Execution:} The path planning and object manipulation module developed in Robot Operating System 1 (ROS) \cite{ros} is responsible for executing the plans generated by the LLM. It translates waypoints or landmarks into motion commands using services such as $navigate\_to\_location$ for path following, $spin\_service$ for rotational scanning, $grab\_object$ for picking up objects, and $drop\_object$ to release them. This configuration enables efficient exploration of the environment while detecting objects of interest with which can interact. Physically gripping objects falls outside of the scope of the system, meaning during manipulation-based tasks a gripping or dropping-like movement is executed near targets to simulate moving an object. 

\textbf{Object Detection and Knowledge Expansion:} The object detector selected for this work is YOLO-v5n \cite{yolov5}, which is a robust, open source, compact, and easy to use object detector compatible with TensorRT \cite{tensorrt} acceleration. TensorRT optimizations such as precision reduction and layer fusion improve inference speed and energy efficiency, making the model suitable for edge deployment. Detected objects are localized using depth sensors and filtered to determine their relevance as landmarks. Relevant objects are added to the KB, which evolves dynamically, enabling the agent to improve its understanding of the environment over time.

\textbf{Edge LLM Deployment:} Deploying an LLM on the memory-constrained Jetson Nano requires balancing computational efficiency and performance. TinyLLama was selected for its ability to deliver competitive performance within a reduced computational footprint. The deployment leverages llama.cpp \cite{gerganov2023llamacpp}, which enables splitting model components between the GPU and CPU swap memory, which is vital when balancing the load of the object detector on the limited memory of the Jetson Nano. The TinyLLama model is fine-tuned and quantized using the Q5\_K medium quantization method, provided by llama.cpp, with the Unsloth \cite{unsloth} framework, ensuring alignment with task-specific scenarios. This approach allows the model to reliably translate natural language prompts into focused actionable plans while minimizing memory and processing requirements. 

\textbf{Dynamic Process Scheduling:} To minimize latency and power consumption, the system employs dynamic scheduling for computationally intensive processes. To ensure objects are not missed when navigating in the environment, the object detector continuously evaluates images while the LLM planner is loaded and only initialized when a new prompt becomes available. This method ensures the system can respond to dynamic changes in the environment while preserving computational and energy resources.

The integration of these components ensures the system can handle hierarchical tasks, such as multi-step navigation and manipulation, in a resource-constrained setting.

The deployment pipeline is summarized as follows:
\begin{itemize}
    \item YOLO asynchronously detects objects and localizes them using depth data, which are filtered and updated into the KB dynamically.
    \item The user provides a natural language task to the LLM to generate a structured plan with waypoints, objects or goals.
    \item Using generated subtasks the agent executes the plan using navigation, scanning, and object manipulation actions.
\end{itemize}

By combining TensorRT-optimized YOLO and GPU-accelerated LLMs with dynamic scheduling, the system achieves a balance between performance and resource efficiency. Comprehensive profiling of the system demonstrates the viability of deploying generative AI on edge devices, showcasing its ability to perform high-level reasoning and real-time decision-making in diverse, real-world scenarios.

%% file: sections/results.tex
\section{Experimental Setup and Results}
This section outlines the experimental setup and results used to evaluate the proposed system and its components in a real world environment when deployed on a robotic agent.

\begin{table*}[bt]
\centering
\begin{tabular}{|c|c|c|cc|}
\hline
\multirow{2}{*}{\textbf{Metric}}                                       & \multirow{2}{*}{\textbf{Input Prompt}}                                                                            & \multirow{2}{*}{\textbf{Cohere}}                                                                                       & \multicolumn{2}{c|}{\textbf{Quantized TinyLlama}}                                                                                                                                                                                                                                                                                                  \\ \cline{4-5} 
                                                                       &                                                                                                                   &                                                                                                                        & \multicolumn{1}{c|}{\textbf{Original (Pre-Trained)}}                                                                                                                                                            & \textbf{Fine-Tuned}                                                                                                    \\ \hline
\textbf{\begin{tabular}[c]{@{}c@{}}Navigation\\ Prompt\end{tabular}}   & \begin{tabular}[c]{@{}c@{}}Navigate to the\\ garage to check\\ if the delivery\\ truck is still here\end{tabular} & \begin{tabular}[c]{@{}c@{}}1/1\\ navigate(garage)\end{tabular}                                                         & \multicolumn{1}{c|}{\begin{tabular}[c]{@{}c@{}}0/1\\ I am not able to\\ perform tasks, but\\ I can provide you with\\ a step-by-step guide\\ on how to navigate ...\end{tabular}}                               & \begin{tabular}[c]{@{}c@{}}1/1\\ navigate(garage)\end{tabular}                                                         \\ \hline
\textbf{\begin{tabular}[c]{@{}c@{}}Manipulation\\ Prompt\end{tabular}} & \begin{tabular}[c]{@{}c@{}}Go to the vending\\ machine grab a\\ bottle and bring\\ it to the office\end{tabular}  & \begin{tabular}[c]{@{}c@{}}4/4\\ navigate(vending –\\ machine)\\ grab(bottle)\\ navigate(office)\\ drop()\end{tabular} & \multicolumn{1}{c|}{\begin{tabular}[c]{@{}c@{}}0/4\\ Here’s a possible\\ solution:1.Navigate to the\\ vending machine 2. grab\\ a bottle from the shelf\\ 3. Bring the bottle\\ to the office ...\end{tabular}} & \begin{tabular}[c]{@{}c@{}}4/4\\ navigate(vending –\\ machine)\\ grab(bottle)\\ navigate(office)\\ drop()\end{tabular} \\ \hline
\textbf{\begin{tabular}[c]{@{}c@{}}Our Prompt\\ Score\end{tabular}}    & \textbf{-}                                                                                                        & \textbf{29/30}                                                                                                         & \multicolumn{1}{c|}{\textbf{0/30}}                                                                                                                                                                              & \textbf{27/30}                                                                                                         \\ \hline
\end{tabular}
\caption{Evaluation of LLM performance, comparing the cloud-based Cohere LLM as a baseline to the pre-trained TinyLlama model and our fine-tuned model post quantization. Results depict the total correctly generated sub-tasks versus the expected sub-tasks across 12 high-level natural language prompts. Two representative sample prompts are included, showcasing the expected sub-tasks alongside the sub-tasks generated by each model.}
\label{tab:llm_comparison}
\end{table*}

\subsection{Robotic Agent and Environment}
The Yahboom Transbot \cite{yahboom_transbot} was selected as the robotic platform (the agent) for executing the desired tasks, which include both navigation and manipulation shown in Figure \ref{fig:environment} (a). The Jetson Nano 4GB edge computing device, which, along with 8 GB of swap memory, manages the autonomous control of the agent and supports the deployment of the \sys{} LLM-guided navigation package.

\sys{} was evaluated in a real-world environment constructed within an office space. The environment included foam pads to emulate various sections of a typical home and office, as depicted in Figure \ref{fig:environment} (b) and (c). The home setting \ref{fig:environment} (b), featured three predefined landmarks in the KB: living room, kitchen, and kids room, while the office setting \ref{fig:environment} (c), contained four landmarks: lounge, lobby, office, and meeting room. Each landmark contains relevant objects to the object detector, for example, the kitchen included items such as a cup, bowl, apple, orange, and bottle, while the living room contained a laptop and keyboard.

In this environment, the Yahboom Transbot was initialized with a pre-loaded map containing the defined landmarks. Using LiDAR-based mapping and simultaneous localization and mapping (SLAM) \cite{Thrun2008} packages provided by ROS, the agent navigated between these landmarks. Additionally, through the integration YOLOv5n the robot identified objects within the environment to incrementally build a KB. This enables the agent to plan and execute future tasks effectively. In between tasks, the agent is controlled remotely via an SSH connection to provide new high-level prompts to the LLM planner for controlling the agent.

\subsection{Edge LLM Fine-Tuning and Compression}
To ensure optimal resource utilization when deploying the TinyLLama LLM after applying quantization, the model is reduced to a size of 745 MB, ensuring compatibility with the Jetson Nano's limited memory. However, since TinyLLaMA is pre-trained with general knowledge, it requires fine-tuning to produce reliable outputs tailored to the specific prompts and tasks for \sys{}.

To illustrate the need for fine-tuning, table \ref{tab:llm_comparison} presents a comparison of TinyLLaMA's original outputs and the cloud LLM Cohere's \cite{cohere2025} responses for sample prompts. While TinyLLaMA generates relevant outputs, they lack the specificity required to decode low-level tasks effectively. For example, two sample prompts and comparison to the baseline cloud LLM can be observed in columns 3 and 4 of table \ref{tab:llm_comparison} showing the TinyLLama LLM provides relevant outputs, however lack focus needed for deployment.


Prior to fine-tuning, a dataset was constructed to expose TinyLLaMA to prompts and required low-level sub-tasks similar to the tasks it will encounter when deployed. \\Two prompt templates (or "skeletons") were created:
\begin{enumerate}
    \item \textbf{Manipulation-based tasks}, involving a sequence of tasks to instruct the agent to navigate to an object and move it to another location.
    \item \textbf{Navigation-based tasks}, for directing the agent to navigate to a specific destination.
\end{enumerate}
An example of how these templates are used by the model is shown in column 3 of Table \ref{tab:llm_comparison}, which displays Coheres' responses to both prompt types.

Using these templates, class names detected by YOLOv5n were inserted, creating an initial dataset. To further expand the dataset and introduce variation, prompts were augmented and reworded using ChatGPT-4. The dataset included diverse system headers with varying KBs to produce context for the LLM creating the final dataset of 20,000 samples.

\begin{figure*} [bt] 
    \centering
    \includegraphics[width=0.98\textwidth]{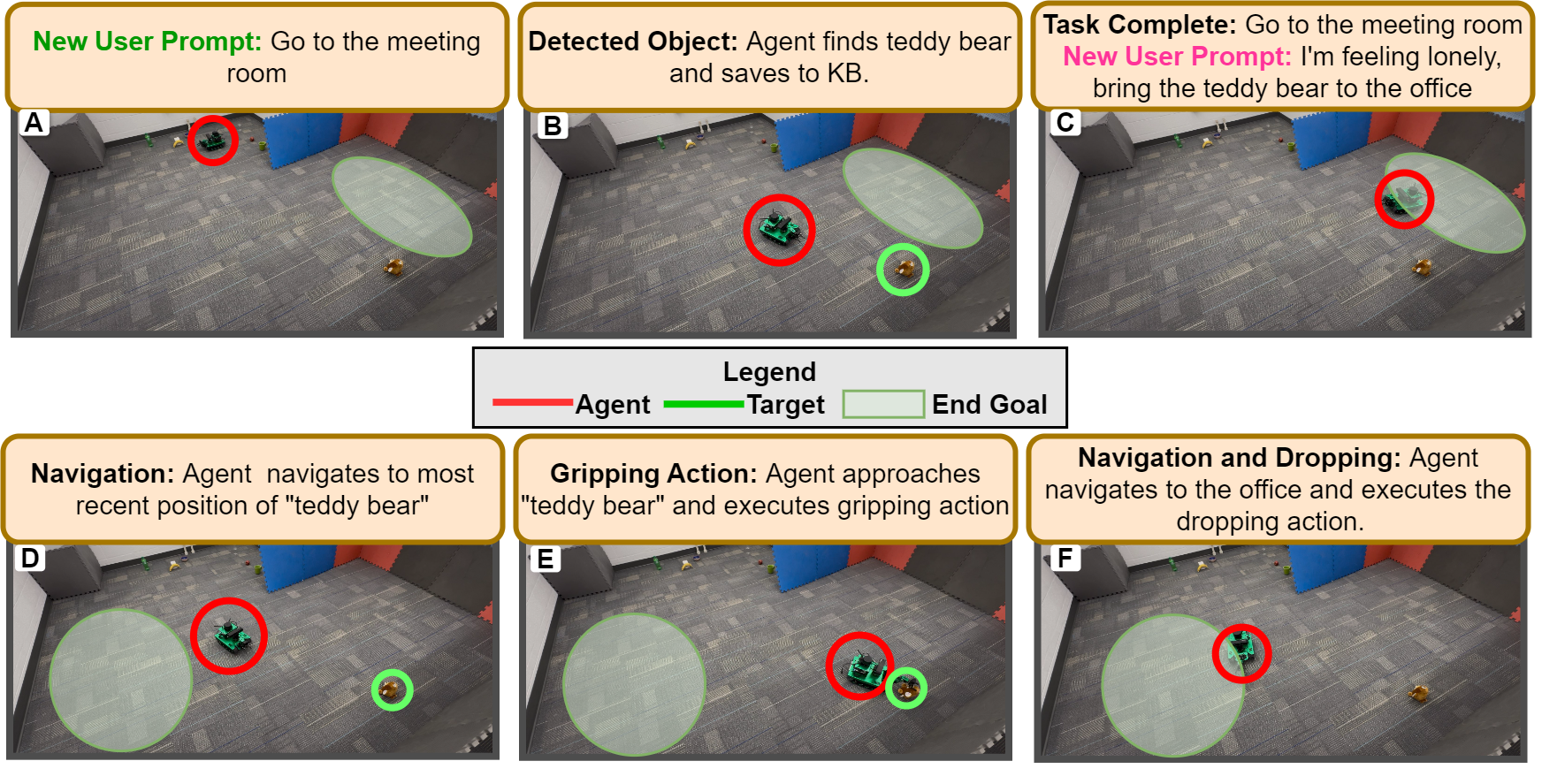}

    \caption{Samples from the office setting experiment showing the agent successfully completing high-level tasks provided to the onboard LLM. The Figure displays two of the prompts executed by the agent: "Go to the meeting room" (A, B, and C) and "I'm feeling lonely, bring the teddy bear to the office" (C, D, E, and F). For the first prompt the agent successfully navigates to the meeting room (C) and adds the "teddy bear" to the knowledge base along its path (B). Then, in the second prompt the agent successfully performs actions representing grabbing the teddy bear (E) and moving it to the office (F).}
    \label{fig:env2_demo}
\end{figure*}

TinyLLaMA was fine-tuned using the generated dataset of 15,000 training and 5,000 testing samples to enable decomposition of natural language prompts into sequential low-level actions. Fine-tuning involved training the model using a batch size of 5, over three epochs with a learning rate of $5\times10^{-5}$. Before training, the model's loss on the dataset was 2.3; after fine-tuning, the loss decreased to 0.4, demonstrating significant improvement.

After fine-tuning, the TinyLLaMA model was quantized, reducing its size to 745 MB. The fine-tuned LLM was then evaluated using natural language prompts to verify its performance. Notably:
\begin{itemize}
\item The LLM accurately responded to prompts containing objects it was fine-tuned on including simple navigation and manipulation tasks.

\item For prompts involving entities (e.g., garage, lounge) not exposed during fine-tuning, the LLM generalized effectively, responding with navigation or manipulation subtasks containing the target entity.

\item Complex prompts containing additional context (e.g., table \ref{tab:llm_comparison} row 2 column 5) were successfully decomposed into actionable outputs.
\end{itemize}
The responses to the original sample prompts provided to the pre-trained TinyLLaMa LLM are displayed in column 5 of table \ref{tab:llm_comparison}. Comparing to the original outputs the fine-tuned LLM provides a focused and accurate response.

These results demonstrate that the fine-tuned TinyLLaMA model is robust to input variations and capable of accurately handling tasks it was not explicitly trained on. This adaptability ensures its effective deployment on the Jetson Nano for real-world applications.

\begin{figure*} [bt] 
    \centering
    \includegraphics[width=\textwidth]{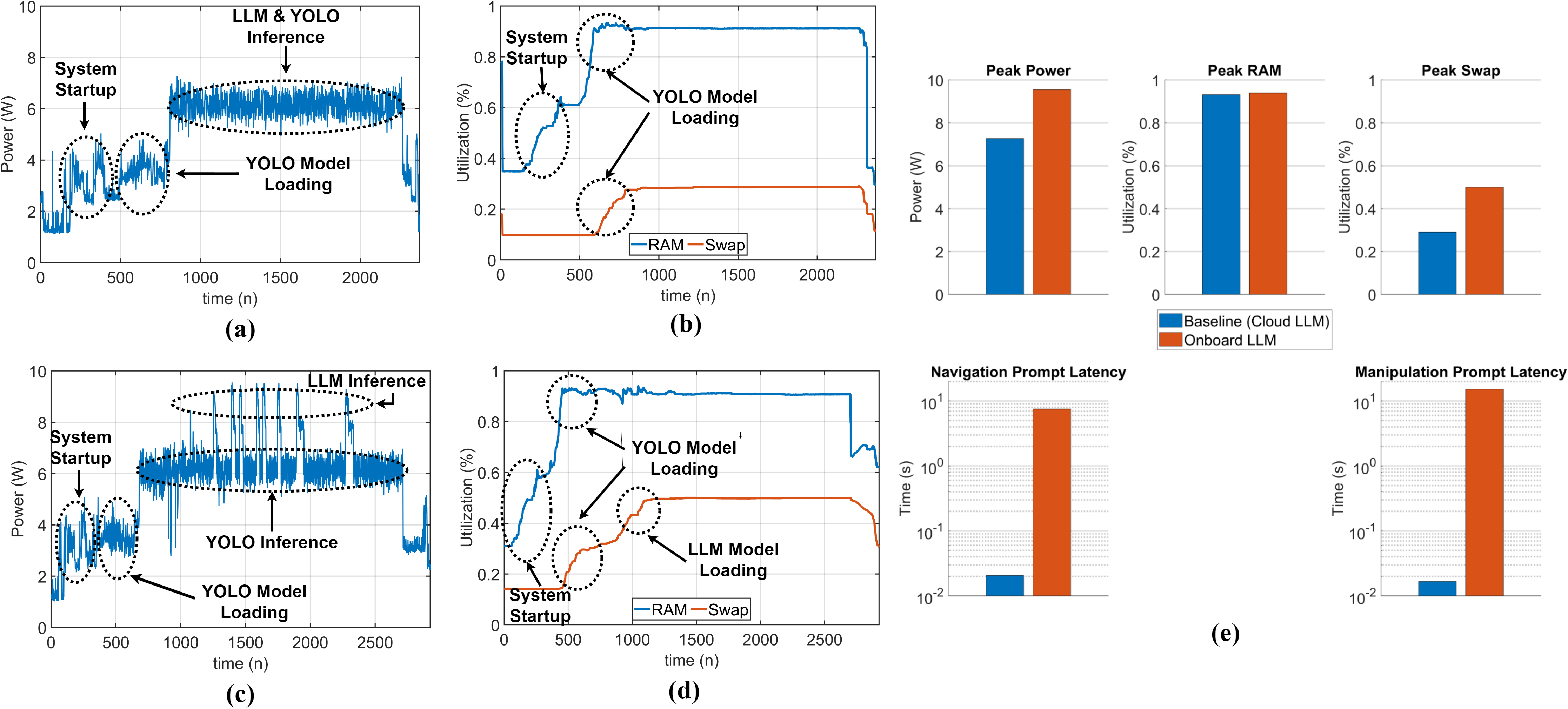}
    \caption{Total power, RAM utilization, swap utilization, and latency results for navigation and manipulation based tasks on the Jetson Nano, run in 10W mode using a 1.43 ghz CPU clock frequency, 921 mhz GPU clock frequency, 4 GB of RAM and 8 GB of swap, during the real-world office setting experiment for the cloud and onboard implementations. The Figure shows results over time for (a) power consumption when deploying the system using the cloud LLM Cohere, (b) memory utilization for the system using the cloud LLM Cohere, (c) power for the fully onboard system (d) memory utilization for the fully onboard system and (e) bar plots for peak power and memory utilization along with prompt processing latency results for simple navigation prompts and more complex manipulation prompts. }
    \label{fig:env2_results}
\end{figure*}

\subsection{Full System Deployment and Evaluation}

With a significant portion of the Jetson Nano's RAM allocated to the object detector, components of the LLM must be distributed between the GPU and CPU swap memory. To address this challenge, the optimized llama.cpp framework is deployed which supports running LLMs on resource-constrained devices like the Jetson Nano by allowing selective offloading of layers to the GPU, while the remaining layers are managed between RAM and swap memory. Although additional latency is produced from CPU processing, this overhead is marginal, as the LLM is queried only at the beginning of a sequence to generate a plan or evaluated during execution of previous plans.

To ensure efficiency and dynamic resource management, the LLM is loaded into memory as a separate process and initialized for decoding only when a new prompt is available. This approach conserves computational resources while maintaining the responsiveness required for high-level task planning.

With the object detector and LLM high-level planner integrated, the agent is deployed in the two emulated home and an office settings described previously. Using an initial map and pre-loaded landmarks, the \sys{} KB building package is initialized alongside the navigation system. The TensorRT-optimized YOLOv5n object detector is launched to analyze images and identify objects in the environment, contributing to the KB. Finally, the LLM is initialized to enable user interaction through natural language prompts. 

The system is evaluated under two configurations:
\begin{itemize}
\item Cloud-based LLM (Cohere): A cloud-based LLM serves as the high-level planner, establishing a baseline for comparison.

\item Fully onboard fine-tuned LLM: The fine-tuned TinyLLaMA model is deployed entirely on the Jetson Nano.
\end{itemize}

A sample demonstration showing the agent performing the desired actions outlined in two sample prompts, in real-time, using the office environment setting and onboard LLM planner is shown in Figure \ref{fig:env2_demo}. Both configurations successfully decomposed high-level tasks into actionable low-level sub-tasks, enabling the agent to navigate the environment and perform actions representing object manipulation tasks, such as gripping and dropping actions near specified targets. 

For both configurations, the agent receives initial prompts to navigate to pre-loaded landmarks, building the KB. Once the KB is established, the agent is tasked with more complex manipulation prompts, such as transporting an object to a new location. Key metrics including LLM prompt decomposition, successful execution, power consumption, RAM usage, and swap utilization are recorded for analysis. The prompts used to evaluate the system in both environments are displayed in column 2 of Table \ref{fig:kb_motivation}. 

\begin{table}[t]
\centering
\begin{tabular}{|c|c|cc|}
\hline
\multirow{2}{*}{\textbf{Setting}} & \multirow{2}{*}{\textbf{Input Prompt}}                                                                                    & \multicolumn{2}{c|}{\textbf{\begin{tabular}[c]{@{}c@{}}Sub-Tasks \\ (\# Success/ Total)\end{tabular}}}                                             \\ \cline{3-4} 
                                  &                                                                                                                           & \multicolumn{1}{c|}{\textbf{\begin{tabular}[c]{@{}c@{}}Fixed\end{tabular}}} & \textbf{\begin{tabular}[c]{@{}c@{}}Growing\end{tabular}} \\ \hline
\multirow{5}{*}{\textbf{Home}}    & \textit{Go to the kitchen}                                                                                                & \multicolumn{1}{c|}{1/1}                                                          & 1/1                                                            \\ \cline{2-4} 
                                  & \textit{Go to the kids room}                                                                                              & \multicolumn{1}{c|}{1/1}                                                          & 1/1                                                            \\ \cline{2-4} 
                                  & \textit{Go to the living room}                                                                                            & \multicolumn{1}{c|}{1/1}                                                          & 1/1                                                            \\ \cline{2-4} 
                                  & \textit{\begin{tabular}[c]{@{}c@{}}I'm hungry bring the \\ banana to the laptop\end{tabular}}                              & \multicolumn{1}{c|}{\textbf{0/4}}                                                 & \textbf{4/4}                                                   \\ \cline{2-4} 
                                  & \textit{\begin{tabular}[c]{@{}c@{}}My son forgot his teddy \\ bear, take the teddy bear \\ to the kids room\end{tabular}}   & \multicolumn{1}{c|}{\textbf{1/4}}                                                 & \textbf{4/4}                                                   \\ \hline
\multirow{6}{*}{\textbf{Office}}  & \textit{Go to the lounge}                                                                                                 & \multicolumn{1}{c|}{1/1}                                                          & 1/1                                                            \\ \cline{2-4} 
                                  & \textit{Go to the lobby}                                                                                                  & \multicolumn{1}{c|}{1/1}                                                          & 1/1                                                            \\ \cline{2-4} 
                                  & \textit{Go to the office}                                                                                                 & \multicolumn{1}{c|}{1/1}                                                          & 1/1                                                            \\ \cline{2-4} 
                                  & \textit{Go to the meeting room}                                                                                           & \multicolumn{1}{c|}{1/1}                                                          & 1/1                                                            \\ \cline{2-4} 
                                  & \textit{\begin{tabular}[c]{@{}c@{}}I'm feeling lonely, bring \\ the teddy bear to the office\end{tabular}}                 & \multicolumn{1}{c|}{\textbf{1/4}}                                                 & \textbf{4/4}                                                   \\ \cline{2-4} 
                                  & \textit{\begin{tabular}[c]{@{}c@{}}Guests are here and they \\are  thirsty bring the\\  bottle to the lobby\end{tabular}} & \multicolumn{1}{c|}{\textbf{1/4}}                                                 & \textbf{4/4}                                                   \\ \hline
\end{tabular}
\caption{Onboard LLM comparison between a fixed KB and our dynamic growing KB for the real world home and office setting experiments. Prompts are provided in sequence for each setting and the number of correctly executed sub-tasks is recorded for the fixed and growing KB. Initial prompts use locations in the initial KB followed by tasks with objects not loaded into the initial KB shown in bold.}
\label{fig:kb_motivation}
\vspace{-10pt}
\end{table}

The system utilization results in Figure \ref{fig:env2_results}, reveal resource utilization during the experiment in the second office environment for cloud-based and on-board configurations:
\begin{itemize}
\item Power Consumption: In both configurations, loading YOLOv5n causes power consumption to stabilize at approximately 6 W. In the cloud-based LLM approach, power consumption remains constant, with no significant change during LLM prompting. In contrast, the fully on-board configuration shows notable spikes to 9.2 W during LLM processing.

\item Memory and Swap Utilization: For both configurations due to the load of the object detector, KB, and navigation packages the RAM utilization remains at a steady 92\% to manage both swap and GPU memory. In contrast, the swap utilization reaches approximately 25\% for the off-board approach, while the LLM configuration on-board doubles the swap to 50\%. 

\item Latency: in the cloud-based approach prompt response latency averaged to 20 milliseconds for all tasks while the onboard approach produced latencies of 8 seconds, given simple navigation-based tasks, and up to 10 seconds, given more complex manipulation-based tasks.
\end{itemize}

These results highlight the computational load and full deployment of \sys{} on the Jetson Nano in a real-world scenario. While the cloud-based configuration offers advantages in terms of prompt response latency and consistent power consumption, it relies on robust network connectivity, which may not always be feasible in dynamic or remote environments. On the other hand, the fully onboard configuration demonstrates a clear trade-off, with increased power and memory demands leading to greater latency. 

To emphasize the benefit of the developing KB, Table \ref{fig:kb_motivation} presents the prompts provided to the LLM planner for both experimental settings alongside the number of successful tasks executed by the agent. The agent is evaluated in a scenario where it cannot grow the KB versus the \sys{} growing KB approach. During these experiments, the agent is first given tasks to navigate to the initial pre-loaded landmarks so the agent begins to explore the environment detecting objects along the way. Then, the agent is provided with more complex manipulation based tasks, using objects which were not available when the system was initialized, to determine the effectiveness of the growing KB.  When the system is unable to grow the KB, the agent fails to complete the manipulation prompts while, leveraging the \sys{} approach to save landmark positions for use in future tasks enables successful task completion. 

%% file: sections/conclusion.tex
\section{Conclusion and Future Work}
This work demonstrates deployment of \sys{} a system architecture integrating generative AI and edge computing for autonomous navigation and task execution in indoor, controlled real-world environments. By integrating a fine-tuned TinyLLM with real-time object detection and efficient path planning, \sys{} achieves dynamic, hierarchical task planning and execution on the resource-constrained Jetson Nano. The approach not only bridges the gap between simulated and real-world deployments but also establishes a framework for integration of a cohesive fully on-board solution capable of executing complex multi-stage tasks.

Future work will focus on evaluating \sys{} in more complex outdoor environments and enhancing the system’s capabilities through the integration of more sophisticated open-vocabulary object detection, advanced object manipulation techniques, and improved environment exploration. These advancements aim to increase the versatility of \sys{} in dynamically changing outdoor environments.

%% file: sections/acknowledgement.tex
\section{Acknowledgments}
We thank Dr. Xiaomin Lin for his review and valuable comments that helped improve this paper. This project was sponsored by the U.S. Army Research Laboratory.

%% file: aaai25.bbl
\begin{thebibliography}{27}
\providecommand{\natexlab}[1]{#1}

\bibitem[{Chen et~al.(2024)}]{chen2024octoplanner}
Chen, Y.; et~al. 2024.
\newblock Octo-planner: On-device Language Model for Planner-Action Agents.
\newblock \emph{arXiv preprint arXiv:2406.18082}.

\bibitem[{Cohere(2025)}]{cohere2025}
Cohere. 2025.
\newblock Cohere - Natural Language Processing Platform.
\newblock Accessed: 2025-01-10.

\bibitem[{Devlin et~al.(2018)}]{DBLP:journals/corr/abs-1810-04805}
Devlin, J.; et~al. 2018.
\newblock {BERT:} Pre-training of Deep Bidirectional Transformers for Language Understanding.
\newblock \emph{CoRR}, abs/1810.04805.

\bibitem[{Dorbala et~al.(2024)}]{dorbala2024llmsgeneratehumanlikewayfinding}
Dorbala, V.~S.; et~al. 2024.
\newblock Can LLMs Generate Human-Like Wayfinding Instructions? Towards Platform-Agnostic Embodied Instruction Synthesis.
\newblock arXiv:2403.11487.

\bibitem[{Gerganov(2023)}]{gerganov2023llamacpp}
Gerganov, G. 2023.
\newblock llama.cpp.
\newblock Accessed: 2025-01-06.

\bibitem[{Gill et~al.(2024)}]{10.1007/s10586-024-04686-y}
Gill, S.~S.; et~al. 2024.
\newblock Edge AI: A Taxonomy, Systematic Review and Future Directions.
\newblock \emph{Cluster Computing}, 28(1).

\bibitem[{Han et~al.(2023)}]{unsloth}
Han, D.; et~al. 2023.
\newblock Unsloth.
\newblock Accessed: 2025-01-06.

\bibitem[{Jocher et~al.(2020)}]{yolov5}
Jocher, G.; et~al. 2020.
\newblock YOLOv5: A state-of-the-art object detection model.
\newblock Accessed: 2025-01-06.

\bibitem[{Kallakuri et~al.(2024)}]{kallakuri2024atlas}
Kallakuri, U.; et~al. 2024.
\newblock {ATLAS}: Adaptive Landmark Acquisition using {LLM}-Guided Navigation.
\newblock In \emph{First Vision and Language for Autonomous Driving and Robotics Workshop}.

\bibitem[{Kam et~al.(2015)}]{10.1007/s11235-015-0034-5}
Kam, H.~R.; et~al. 2015.
\newblock RViz: a toolkit for real domain data visualization.
\newblock \emph{Telecommun. Syst.}, 60(2): 337–345.

\bibitem[{Lin et~al.(2024)}]{10543121}
Lin, B.; et~al. 2024.
\newblock Correctable Landmark Discovery via Large Models for Vision-Language Navigation.
\newblock \emph{IEEE Transactions on Pattern Analysis and Machine Intelligence}, 46(12): 8534--8548.

\bibitem[{Mazumder and Mohsenin(2023)}]{mazumder2023reg}
Mazumder, A.~N.; and Mohsenin, T. 2023.
\newblock Reg-TuneV2: Hardware-Aware and Multi-Objective Regression-Based Fine-Tuning Approach for DNNs on Embedded Platforms.
\newblock \emph{IEEE Micro}.

\bibitem[{Navardi et~al.(2023)}]{micro2023-mozhgan}
Navardi, M.; et~al. 2023.
\newblock Metae2rl: Toward metareasoning for energy-efficient multi-goal reinforcement learning with squeezed edge yolo.
\newblock \emph{IEEE Micro}.

\bibitem[{Navardi et~al.(2024)}]{10.1109/LES.2024.3446948}
Navardi, M.; et~al. 2024.
\newblock MetaTinyML: End-to-End Metareasoning Framework for TinyML Platforms.
\newblock \emph{IEEE Embed. Syst. Lett.}, 16(4).

\bibitem[{NVIDIA(2023)}]{tensorrt}
NVIDIA. 2023.
\newblock TensorRT: A high-performance deep learning inference library.
\newblock Accessed: 2025-01-06.

\bibitem[{Prakash et~al.(2024{\natexlab{a}})}]{prakash2024using}
Prakash, B.; et~al. 2024{\natexlab{a}}.
\newblock Using LLMs for Augmenting Hierarchical Agents with Common Sense Priors.
\newblock In \emph{The International FLAIRS Conference Proceedings}, volume~37.

\bibitem[{Prakash et~al.(2024{\natexlab{b}})}]{bharat2024using}
Prakash, B.; et~al. 2024{\natexlab{b}}.
\newblock Using LLMs for Augmenting Hierarchical Agents with Common Sense Priors.
\newblock In \emph{The International FLAIRS Conference Proceedings}, volume~37.

\bibitem[{Rajvanshi et~al.(2024)}]{rajvanshi2024saynavgroundinglargelanguage}
Rajvanshi, A.; et~al. 2024.
\newblock SayNav: Grounding Large Language Models for Dynamic Planning to Navigation in New Environments.
\newblock arXiv:2309.04077.

\bibitem[{Shah et~al.(2023)}]{pmlr-v205-shah23b}
Shah, D.; et~al. 2023.
\newblock Lm-nav: Robotic navigation with large pre-trained models of language, vision, and action.
\newblock In \emph{Conference on robot learning}, 492--504. PMLR.

\bibitem[{Shaharear et~al.(2024)}]{ragib2024vitreg}
Shaharear, M.~R.; et~al. 2024.
\newblock ViT-Reg: Regression-Focused Hardware-Aware Fine-Tuning for ViT on tinyML Platforms.
\newblock \emph{IEEE Design \& Test}.

\bibitem[{Singh et~al.(2023)}]{SINGH202371}
Singh, R.; et~al. 2023.
\newblock Edge AI: A survey.
\newblock \emph{Internet of Things and Cyber-Physical Systems}, 3: 71--92.

\bibitem[{Song et~al.(2023)}]{song2023llmplannerfewshotgroundedplanning}
Song, C.~H.; et~al. 2023.
\newblock LLM-Planner: Few-Shot Grounded Planning for Embodied Agents with Large Language Models.
\newblock arXiv:2212.04088.

\bibitem[{{Stanford Artificial Intelligence Laboratory et al.}(2018)}]{ros}
{Stanford Artificial Intelligence Laboratory et al.} 2018.
\newblock Robotic Operating System.

\bibitem[{Thrun and Leonard(2008)}]{Thrun2008}
Thrun, S.; and Leonard, J.~J. 2008.
\newblock \emph{Simultaneous Localization and Mapping}, 871--889.
\newblock Springer Berlin Heidelberg.
\newblock ISBN 978-3-540-30301-5.

\bibitem[{Yahboom(2025)}]{yahboom_transbot}
Yahboom. 2025.
\newblock Transbot for Jetson Nano.
\newblock Accessed: 2025-01-06.

\bibitem[{Zhang et~al.(2024)}]{zhang2024tinyllama}
Zhang, P.; et~al. 2024.
\newblock TinyLlama: An Open-Source Small Language Model.
\newblock Technical Report, arXiv:2401.02385.

\bibitem[{Zhu et~al.(2024)}]{10734363}
Zhu, Y.; et~al. 2024.
\newblock ChatNav: Leveraging LLM to Zero-shot Semantic Reasoning in Object Navigation.
\newblock \emph{IEEE Transactions on Circuits and Systems for Video Technology}, 1--1.

\end{thebibliography}
